# Multi-Sensor Terrestrial SLAM for Real-Time, Large-Scale, and GNSS-Interrupted Forest Mapping

Weria Khaksar
Faculty of Science and Technology, Norwegian University of Life Sciences (NMBU), Ås, Norway. weria.khaksar@nmbu.no

Rasmus Astrup
Division of Forest and Forest Resources, Norwegian Institute of Bioeconomy Research (NIBIO), Ås, Norway.
rasmus.astrup@nibio.no

**Abstract:**

Forests, as critical components of our ecosystem, demand effective monitoring and management. However, conducting real-time forest inventory in large-scale and GNSS-interrupted forest environments has long been a formidable challenge. In this paper, we present a novel solution that leverages robotics and sensor-fusion technologies to overcome these challenges and enable real-time forest inventory with higher accuracy and efficiency. The proposed solution consists of a new SLAM algorithm to create an accurate 3D map of large-scale forest stands with detailed estimation about the number of trees and the corresponding DBH, solely with the consecutive scans of a 3D lidar and an imu. This method utilized a hierarchical unsupervised clustering algorithm to detect the trees and measure the DBH from the lidar point cloud. The algorithm can run simultaneously as the data is being recorded or afterwards on the recorded dataset. Furthermore, due to the proposed fast feature extraction and transform estimation modules, the recorded data can be fed to the SLAM with higher frequency than common SLAM algorithms. The performance of the proposed solution was tested through filed data collection with hand-held sensor platform as well as a mobile forestry robot. The accuracy of the results was also compared to the state-of-the-art SLAM solutions.

## 1. Introduction:

Forests are among the most important players in the climate change and global carbon cycle as they cover one third of the global land area (Köhl et al. 2015). This vital role of forests has attracted researchers and policymakers to increase accuracy and productivity in the development of forest resource information systems (Trumbore et al. 2015). Such systems are needed for forest development status evaluation, such as climate change impact evaluations and carbon stock estimations, and forestry departments use it as the basis for creating forest development plans, such as forest cultivation and forest harvesting plans (Fan et al. 2020; Nesha et al. 2021). The accurate spatial distribution and attributes (height, diameter at breast height (DBH), species, etc.) of trees are among the most important elements of forest inventories, decision making on forest resources, and ecological studies.

Evidently, it is expensive to measure sample plots with conventional measurement tools (Austin et al. 2020). With the rapid development and improvement of sensors (Bayne et al. 2017) and robotics (Oliveira et al. 2021) in forestry, the implementation of these technology in forestry is getting widely adopted to obtain more accurate and low-cost forest inventory. A commonly used sensor for forest management is known as light detection and ranging (LIDAR) which has been implemented in aerial or terrestrial approaches (LaRue et al. 2020). Besides the conventional standalone Terrestrial Laser Scanning (TLS) techniques, various type of lidar sensors have been attached to moving platforms such as Unmanned Aerial Vehicles (UAV) and Unmanned Ground Vehicles (UGV) for forest inventory remote sensing on larger scales.

Despite numerous advantages of mobile laser scanning in forests, a main challenge is to overcome the positional accuracy which is mainly due to the lack of accurate estimation on the position and orientation of the sensor base as it moves in forest environments (Shao et al. 2021). A common solution for different positioning problems is to use Global Navigation Satellite Systems (GNSS) (Kukko et al. 2012; Holopainen et al. 2013; Qian et al. 2017). However, recent high-precision differential GPS (DGPS) even with Real-Time Kinematic (RTK) capabilities, suffer major signal loss in forest environments where dense canopy absorbs, reflects, or even blocks the signal and causes great error in positioning. Such degraded positioning results greatly reduces the system capability of estimating the exact location of the sensor base. As an attempt to overcome this issue, Inertial Measurement Units (IMU) have been implemented on mobile TLS systems in forests (Cahyadi and Rwabudandi 2019; Aguiar et al. 2020). Despite the advantages of using IMUs, the dependance on a detailed terrain model is a major drawback. Specially in forests, where wheel spinning and slipping makes the odometry almost impossible (Chen et al. 2020). Even with a high-precision GNSS-IMU system, the positioning error can grow to several tens of centimeters and even to meters due to trajectory drift and will greatly limit the mapping accuracy in the dense forest scenes (Shao et al. 2021).

Simultaneous Localization and Mapping (SLAM) is the process of generating a map of an unknown environment while continuously estimating the sensor's location using different modalities of sensors and sources of input. Recently, SLAM has found to be a highly useful tool in forest remote sensing. The main two classes of SLAM, i.e., Visual SLAM using camera, and Lidar-based SLAM, have been adopted to deal with environmental challenges of forests (Gollob et al. 2020; Ali et al. 2020). Despite the tremendous progress made in SLAM in the past 30 years, it is still considered to be an unsolved problem. SLAM is an estimation problem and given the complexities of the environment and the uncertainties in the sensor measurements, it is yet to arrive a complete solution. A good SLAM solution significantly relies on the environment, the robot or the data collection agent, the uncertainties in the sensor measurements and the level of performance that we intended to achieve (Singandhupe and La 2019). Accordingly, it can be concluded that the SLAM problem is highly case-dependent and for each individual application, additional challenges rise which solving them would be critical to achieve a desirable outcome.

In forestry applications, several types of SLAM approaches have been introduced but several factors will affect the performance including forest type, season, soil type, etc. Furthermore, SLAM on its own does not provide sufficient information for forest inventory purposes. A fundamental challenge in forestry SLAM systems is to the design of metric and semantic representations for the environment. Even though the interaction with the environment is paramount for most applications of robotics, most of SLAM systems are not able to provide a tightly coupled high-level understanding of the geometry and the semantic of the surrounding world (Cadena et al. 2016). the design of such representations must be task-driven and currently a tractable framework to link task to optimal representations is lacking.

In this paper, a new SLAM algorithm is proposed to create an accurate 3D map of large-scale forest stands with detailed estimation about the number of trees and the corresponding DBH, solely with the consecutive scans of a 3D lidar and an imu. This method utilized a hierarchical unsupervised clustering algorithm to detect the trees and measure the DBH from the lidar point cloud. The algorithm can run simultaneously as the data is being recorded or afterwards on the recorded dataset. Furthermore, due to the proposed fast feature extraction and transform estimation modules, the recorded data can be fed to the SLAM with higher frequency than common SLAM algorithms.

## 2. The SLAM Problem:

Simultaneous localization and mapping (SLAM) problem comprises the simultaneous estimation of the state of an agent and the construction of a model (map) of the environment that the agent senses. In simple instances, the agent state is described by its pose (position and orientation). The map, on the other hand, is a representation of position of landmarks and obstacles describing the environment in which the agent operates. The need to use a map of the environment is twofold. First, the map is often required to support other tasks; for instance, a map can inform path planning or provide an intuitive visualization for a human operator. Second, the map allows limiting the error committed in estimating the state of the agent (Cadena et al. 2016). In the absence of a map, dead-reckoning would quickly drift over time; on the other hand, using a map, the agent can "reset" its localization error by revisiting known areas (so-called loop closure). Therefore, SLAM finds applications in all scenarios in which a prior map is not available and needs to be built.

A simple formulation of the SLAM problem can be defined with four initial variables:

$P_T = \{p_0, p_1, \ldots, p_T\}$

$U_T = \{u_0, u_1, \ldots, u_T\}$

$$S_T = \{s_0, s_1, \ldots, s_T\}$$

$$M_T = \{m_0, m_1, \ldots, m_{T-1}\}$$

Where $p_t$ is the pose of the sensor (lidar) at time $t$, $u_t$ is the sensor's motion between time $t-1$ and time $t$, $s_t$ is the sensor's measurement at time $t$, and $m_t$ is a vector representing the position of landmarks at time $t$. The SLAM problem can be defined as the problem of recovering a model of the map $M$ and the sensor's path $P_T$, from the odometry $U_T$ and the observations $S_T$ as an estimation problem (Siegwart et al. 2011):

$$P(P_T, M | S_T, U_T)$$

In forestry applications, the SLAM problem becomes even more challenging due to the environmental factors. The unexpected motion of the mapping agent in forests makes the estimation of $U_T$ more difficult. Furthermore, a simple SLAM system might not be sufficient for forestry applications and further information about tree location, dimension, estimated volume, and species are required alongside the core resulted map.

## 3. Related Work:

In almost all forestry tasks, one key element is the availability of tree maps, together with methods enabling the identification and selection of individual trees. In addition, local route planning requires accurate updates on the position of the harvester inside the forest in real-time, which cannot be relied on global navigation satellite system (GNSS) sensors (Fang Liao et al. 2016; Tian et al. 2020). Furthermore, the need for automatic machines in forestry is increasing since farmers increasingly recognize its impact in agriculture (Bergerman et al. 2016). Robots are now used for a variety of tasks such as planting, harvesting, environmental monitoring, supply of water and nutrients, and others (Roldán et al. 2018). In this context, developing solutions that allow robots to navigate safely in these environments is essential. These advances impose a main requirement: localizing the robots in these agriculture and forestry environments. The most common solution is to use Global Navigation Satellite System (GNSS) standalone-based solutions (Perez-Ruiz and K. 2012; Guo et al. 2018). However, in many agricultural and forestry places, satellite signals suffer from signal blockage and multi-reflection (Aguiar et al. 2020), making the use of GNSS unreliable. In this context, it is extremely important to research and develop intelligent solutions that use different modalities of sensors, as well as different sources of input data, to compute the robot localization.

Simultaneous localization and mapping (SLAM) (Cadena et al. 2016) is the state-of-the art approach to do so. This technique consists of estimating the state of an agent using input sensor data, while simultaneously building a map of the surrounding environment. The classical SLAM approaches are based on extended Kalman filters (EKF) (Huang et al. 2008 - 2008), Rao–Blackwellized particle filters (Grisetti et al. 2007), and maximum likelihood estimation (Rybski et al. 2004). Next generation of SLAM systems were more focused on the fundamental properties of SLAM, including observability, convergence, and consistency (Dissanayake et al. 2011 - 2011). A more recent review on SLAM systems is given in (Cadena et al. 2016).

Among numerous SLAM systems currently available, it is worthy to mention few important approaches. A low-drift and real-time lidar odometry and mapping (LOAM) method is proposed (Zhang and Singh 2014, 2017) which performs point feature to edge and plane scan-matching to find correspondences between scans. Features are extracted by calculating the roughness of a point in its local region. The points with high roughness values are selected as edge features. Similarly, the points with low roughness values are designated planar features. Despite superior performances, the performance of LOAM deteriorates when sensing resources are incomplete. Due to the need to compute the roughness of every point in a dense 3D point cloud, the update frequency of feature extraction on a lightweight embedded system cannot always keep up with the sensor update frequency. Furthermore, operation in noisy environments also poses challenges for LOAM. Especially for forestry robots where the mounting position of the lidar is often close to the ground, sensor noise from the ground may be a constant presence (Shan and Englot). A lightweight and ground-optimized lidar odometry and mapping (LeGO-LOAM) method is proposed (Shan and Englot 2018) for mapping tasks of ground rovers. Its fusion of IMU measurements is the same as LOAM. Since LeGO-LOAM maintains the local consistence of the ground plane between consecutive frames, the estimated trajectory is usually flatter than the one estimated by LOAM (Koide et al. 2019a). However, it still suffers from the accumulated rotational error due to the lack of the global ground constraint. Consequently, it fails to estimate the trajectory the motion of the lidar is sudden, frequent, and changing which is highly likely to happen in forestry applications.

The hdl-graph-slam algorithm (Koide et al. 2019a) was introduced as a graph SLAM method that builds a pose graph which is refined with loop closures, and optionally, GPS data. This SLAM system does not use the IMU data for odometry estimation. Odometry is computed from the LiDAR data directly by computing the transformation required to overlap recent scans. Hdl-graph-slam allows the use of IMU data to periodically improve the estimated trajectory by aligning the IMU acceleration vector of each trajectory node with the gravity vector. This approach is computationally heavier that the former slam algorithms as it processes the lidar data as the main source of transformation calculation which makes it less useable for real-time implementation. However, an advantage of this algorithm is its flexibility with slam parameters for offline improvement (Koide et al. 2021). A LiDAR-SLAM system (LIO-SAM) is proposed (Shan et al. 2020 - 2021), in which a factor graph is applied to incorporate the 3-DoF GNSS positioning data as factors into the system. LIO-SAM requires the roll-pitch-yaw estimate from the IMU's built in estimator, which has significant drift due to high-frequency vibration when the sensor is moving in frequently changing environment like forests (Nguyen et al. 2020; He et al. 2021).

The main goals for robotics in forest environments are stem mapping, harvesting, transportation, and biomass estimation (Aguiar et al. 2020). To achieve acceptable performance, SLAM systems has been adapted and further developed to deal with specific requirements of forestry. The use of SLAM has been explored previously in forest environments using 2D LiDAR combined with GPS. The method combines 2D laser localization and mapping with GPS information to form global tree maps. Besides the disadvantages of using 2D data and low accuracy of the results, this method has reported to run slower than it is required for real-time applications (Miettinen et al. 2007). A feasibility study for the utilization of LiDAR-based SLAM on a vehicle platform was conducted (Öhman et al. 2008); however, the accuracy was not provided. A mapping approach is reported (Hyyti and Visala 2013) where the stem diameters are measured, and a ground model is built. This work shows a trunk diameter error of 4 cm for short range observations of less than 8 m while the geometry of the self-built 3D scanner is used as an advantage in the tree trunk and ground detection. In another report (Hussein et al. 2015), a method was suggested to build the global map using aerial ortho-imagery, which enables the navigation with prior map information without mandatory previous visits to the working environment. Using scan-matching concepts, a SLAM-aided positioning solution with point clouds collected by a small-footprint LiDAR was proposed (Tang et al. 2015) based on a 2D SLAM algorithm that solves the Improved maximum likelihood estimation method to localize the robot, with the goal of estimating the biomass forest inventory. In order to build accurate stem maps, GNSS/INS fused with a scan-matching approach was proposed (Qian et al. 2017) based on an improved maximum likelihood estimation method. This work presented an accuracy of 13 cm in a real experiment with 800 m, presenting scalability and availability, since it does not require prior mapping information. Aiming to perform 3D mapping of a forest environment, a graph-based SLAM approach was proposed (Pierzchała et al. 2018) that computes 6-DoF robot pose using the Levenberg–Marquardt algorithm. In addition, this work presents other contributions, such as built in-house datasets that were used to test and evaluate the SLAM pipeline. An end-to-end pipeline for tree diameter estimation has been reported (Chen et al. 2020) based on semantic segmentation and lidar odometry and mapping. This method utilizes semantic feature-based pose optimization that simultaneously refines the tree models while estimating the robot pose. Furthermore, (Li et al. 2020) propose an innovative localization approach based on map matching with Delaunay triangulation (Liu and Yin 2020). Here, the Delaunay triangulation concept is used to match local observations with previously mapped tree trunks. Unlike most works that use point cloud-based matching approaches, this method utilized a topology-based approach and achieved an accuracy of 12 cm in a long-term path with 200 m. On the negative side, it requires a previously built map so that it can operate. Finally, a two-phase on-board process was proposed (Nevalainen et al. 2020), where tree stem registration produces a sparse point cloud (PC) which is then used for SLAM. This method uses only tree stem centers, reducing the allocated memory to approximately 1% of the total PC size.

It can be concluded from the above studies that there is a huge dependency on GNSS-free localization systems from autonomous mobile robots working in forestry and in under-canopy GNSS-interrupted environments, localization is an unsolved challenge. Also, most of the available approaches use clustering-based methods for tree stem detection and modelling, which is limited to relatively clean forests. Another disadvantage of the current proposals is that a small error in the orientation estimation can significantly affect the mapping of trees that are farther away from the sensors which is a serious challenge in large-scale forest mapping. Furthermore, majority of the available systems are not fast enough to be implemented in real-time SLAM applications and a post-processing phase is inevitable for creating a detailed map and tree stem models.

## 4. Proposed Method:

The proposed system receives data from a 3D lidar, an IMU, and a camera. The camera data is only used for online visualization and referencing. The aim of the model is to predict the sensor's state and trajectory based on consecutive sensor observations. This estimation can be viewed as a maximum a posteriori (MAP) estimation problem (Tronarp et al. 2021) in which the objective is to estimate the agent's state, trajectory and environment landmarks. The model is given a set of observation $M = \{m_k : k = 1, \ldots, m\}$ such that each observation can be expressed as a function of F, i.e., $f_k = h_k(f_k) + \epsilon_k$, where $f_k \subseteq F$ is a subset of the variables, $h_k(.)$ is a known function (the measurement or observation model), and $\epsilon_k$ is random measurement noise (Cadena et al. 2016).

As an accepted terminology by the robotics community (Wang et al. 2021), the architecture of the proposed SLAM algorithm is defined by two main components: the front-end and the back-end. The front-end classifies and abstracts the sensor data into different models that are capable of being used for estimation, while the back-end acts as an inference on the data fused by the front-end. An overview of the proposed method is shown in Figure 1.

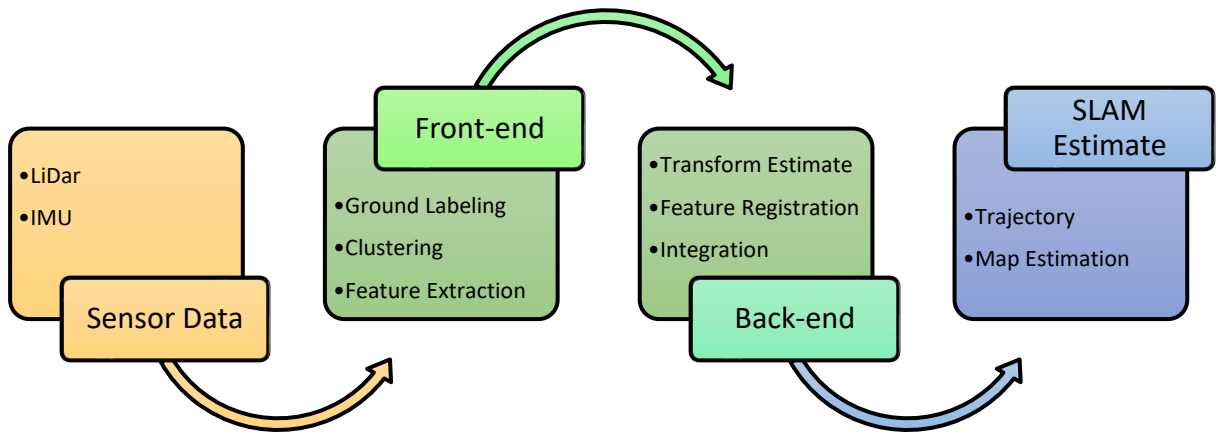

Figure 1. Overview of the proposed SLAM.

The system receives data from a 3D lidar and an imu and outputs estimations on the trajectory and the global map. There are six main components of the system which work in a consecutive fashion to reach the desired output. After receiving the first batch of input data, the ground labeling component starts by estimating the local ground plane. Next, the point cloud clustering module takes place by clustering a single scan point cloud through an unsupervised clustering approach. The clustered point cloud is then processed by the feature extraction model which separates and labels edge points and surface points. The extracted features are then used by the transform estimation module to compute the right transform between two consecutive lidar scans. The extracted features are also processed by the feature registration module which registers current points to a global point cloud structure. Finally, the resulting estimated transforms and registered points are fused into the integration module to compute the final pose estimation. A detailed description of the system components is given below.

## 4.1. Ground Labeling:

Cloud points that represent the ground surface are usually a large part of the point cloud and detecting/labeling them significantly reduces the size of the SLAM problem. To detect the ground points, a local labeling method is adapted from (Rummelhard et al. 2017 - 2017) which performs an adaptive local detection. Other available methods such as (Himmelsbach et al. 2010; Bogoslavskyi and Stachniss 2016) use two-dimensional line extraction approach which fails in curved terrains. The ground surface is modeled as a regular two-dimensional lattice on the *XY* plane. The lattice consists of a set of nodes, each represented by $N_i = (n_i^x, n_i^y)$ and random variable $\Omega_i = (h_i, s_i^x, s_i^y)$ where $h_i$ and $(s_i^x, s_i^y)$ represent the estimated elevation and two-directional slopes of the ground respectively. Each coordinate of raw points in each scan $(x_i, y_i, z_i) \in \mathbb{R}^3$ is associated with the closest node and a binary random labeling variable is defined for each point as $G_i \in \{0,1\}$, indicating the point belongs to the ground ($G_i = 1$) or not ($G_i = 0$). The ground labeling problem is then formulized as estimation of the unknown variables $\Omega_i^t$ and $G_i^t$ with respect to the point cloud coordinates and the random distribution of $\Omega$ at the previous time step $\Omega_i^{t-1}$ (Rummelhard et al. 2017 − 2017). The results of the ground labeling module can even be further used for road detection and assessment purposes (Benyamin et al. 2017; Buján et al. 2021).

## 4.2. Clustering:

After identifying and labeling the ground points, the remaining points with the ground value of zero, $G_i = 0$, need to be categorized into small clusters for further post-processing purposes. The aim of the clustering module is to find a value ($c_i$) for each point cloud member, $(x_i, y_i, z_i) \in \mathbb{R}^3$, that represent its cluster details. Unlike several SLAM algorithms which transform the point cloud into a range image(Zermas et al. 2017; Krispel et al. 2020), the proposed method applies the clustering directly on the 3D input data. Although it would slightly increase the computational cost of the module, the resulted clusters lower the cost of feature extraction module significantly. More importantly, performing direct 3D point cloud clustering enables the intended real-time tree detection and analysis capability of the proposed system. A hierarchical unsupervised clustering algorithm was adapted for point cloud clustering (Cohen-Addad et al. op. 2108). Since the number of clusters is unknown and pretraining the clustering algorithm is not possible, the hierarchical $k$-means clustering algorithm (Murtagh and Contreras 2017) is used for identifying each point with three labels including the cluster index ($c_i$), the range value ($r_i$), and the number of members of the same cluster ($m_i$). These values will be used in the feature extraction module. In general, the implemented clustering method takes place by creating a feature vector, **v** where for each cluster ($j$) at the current stage of the algorithm:

$$v_j = \max(0, p_j)$$

$$p_j = \frac{1}{|C|} \left[ \sum_{k=1}^{|C|} \|i - C_k\|^2 \right] - \|i - C_k\|^2$$

In other words, each cluster is given a descriptor which indicated the relative activation of that cluster. The implemented clustering approach starts by defining a branching factor ($k$) which defines the number of clusters at each level of the clustering hierarchy. At each stage of the algorithm, the standard $k$-means clustering algorithm takes place by receiving the branching factor as the number of clusters. Then, each sub-cluster keeps being recursively clustered until the algorithm reaches a termination criterion which usually is when the number of points in each cluster reaches a predefined value ($\sigma$). We found $\sigma \in [50, 100]$ most suitable during the execution. If there are not enough data points in a given cluster then we end up generating degenerate centroids that will never be activated. While degenerate centroids do not directly hurt classification performance, they do affect the tree detection accuracy. Therefore, it makes sense to maintain a relatively uniform number of data points for each centroid. In hierarchical clustering, it is easily possible to merge clusters with few points into neighboring clusters.

Another advantage of using this clustering technique is to segment and cluster the final resulting point cloud for three detection and DBH estimation. Since the ground points have already been detected, trees are detected using the same hierarchical $k$-means clustering algorithm based on the number of points per cluster and a neighborhood function. To extract the DBH values of each clustered tree, we chose a normalized height value in the range $1.2–1.4\ (m)$, as this represented the standard measurement height of DBH. The points within this range were then projected onto a $2D$ plane for further processing. Before estimating DBH values, outlier points should be identified and removed as they complicate the estimation process and might even cause point cloud registration failure. By computing the mean ($\mu$) and standard deviation ($\sigma$) of $k$ nearest neighbor distances for each point, outliers were identified as points that fall outside the ($\mu \pm \omega.\sigma$) interval. The value of $\omega$ depends on the size of the analyzed neighborhood. In our implementation, we set $\omega = 0.4$ and $k = 10$, since experiments have confirmed the applicability of this threshold, with approximatively 1.4% of the points being outliers. Figure 2 shows the process of DBH estimation where the segmented tree trunk goes through a vertical range filter and after 2D projection and outlier selection, DBH is estimated by fitting a circle to the remaining points.

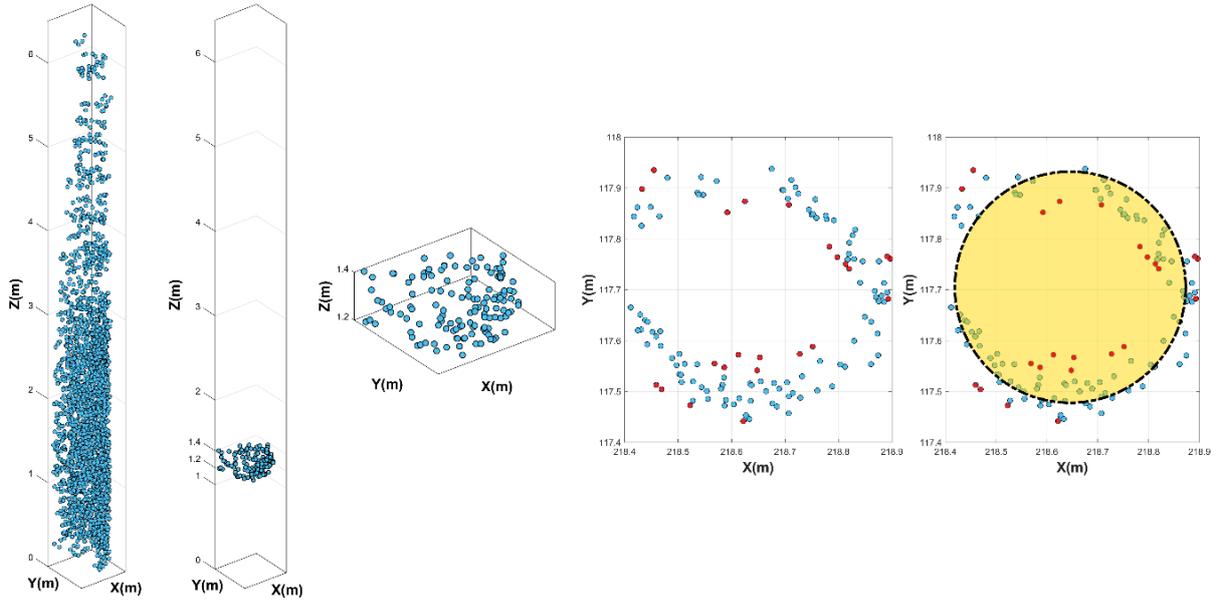

Figure 2. The process of DBH estimation in the clustering module. From left, segmented tree, implementation of the vertical range filter, resulted points, 2D projection with outlier selection, and finally, circle fitting and DBH estimation.

*4.3. Feature Extraction*:

The input from the clustering module will be updated with the IMU readings before feature extraction. As discussed before, a main challenge in forestry SLAM is the quick velocity variations of the mapping agent. To deal with fast velocity changes, the clustered point cloud is corrected in two ways before sending to the feature extraction. First, with orientation from the IMU, each frame is rotated to align with the initial orientation of the lidar in that frame. Second, with acceleration measurement, the motion distortion is partially removed and a constant velocity during the sweep will be assumable. Here, the IMU orientation is obtained by integrating angular rates from gyros and readings from accelerometers in a Kalman filter as implemented in (Zhang and Singh 2017). For convenience, we assume that the IMU frame coincides with the mapping agent, i.e., robot or the handheld device.

The feature extraction module is based on the suggested method used in LOAM with a difference in the feature extraction source. The original method takes the entire point cloud and selects features points in two groups namely, sharp edge, and planar surface. Since the lidar returns points in clockwise or counterclockwise order, any set of consecutive points contains half of its points on each side of a given reference point with a fixed interval ($\sigma$) between two points. The value of $\sigma$ can be calculated based on the resolution settings of the lidar. Then, a decision factor is calculated based on the smoothness of each point and then points will be sorted based on this value where points with the highest values are selected as edge feature points and points with lowest values are selected as planar feature points. In this paper, a different source is used for feature extraction. Unlike the above-mentioned process, the proposed algorithm calculates the decision factor only for the clustered points from the previous modules i.e., ground points, and clustered points. This idea is inspired by the feature extraction process proposed in (Shan and Englot) to make the process faster for real-time implementations. We define our decision factor as:

$$\varphi_i = \frac{|W|.\|r_i\|}{\sum_{j \in W, j \neq i}\|r_i - r_j\|}$$

where $W$ is the set of consecutive points of point $i$ from the same cluster, and $r_j$ is the distance of point $j$ from the center of the lidar. After sorting the points of each scan based on their computed factor, edge and planar feature points are extracted based on maximum and minimum values, respectively.

*4.4. Transform Estimate*:

The role of the transform estimation module is to find the right transform between two consecutive point clouds registered by two consecutive sweeps of the lidar. In other words, it computes the lidar motion between two complete 360º scans. This is based on the lidar used in our experiments and a complete scan could cover less than 360º if a different lidar is used with a limited horizontal view. This module uses the features extracted by the previous module and finds the closest matches for planar and edge points. The closest neighbor point of each new feature point is found in the previous point cloud. Since we are handling edge and planar feature points separately, the right lidar motion can be calculated by minimizing the overall distance of the feature points. By defining separate stacks of edge and planar feature points, the problem can be solved using the Levenberg-Marquardt method (Andrew 2001). Using this method for computing the lidar motion has been suggested and explained in (Zhang and Singh 2017).

*4.5. Feature Registration*:

The output of the transform estimation module will be used alongside the extracted feature points as the input of the feature registration module. This module registers the cloud points into a global map which acts as a combining agent for different point clouds from different full 360º scans. To paste two consecutive point clouds into this global structure, we need to have the feature points from each cloud as well as the correct transform of the lidar between these scans which has been computed in the last two modules. This module is adapted from (Zhang and Singh 2017) and (Shan and Englot). However, we introduce a new structure to index the points. Instead of saving a single point cloud (Zhang and Singh 2017) or saving each individual feature point set (Shan and Englot), a two-stage process is proposed to save the computational cost of the feature registration module and increase accuracy. In the first stage, the feature points are stored in a 20 (m) cubic map and the decision factor as defined before, is used to distinguish planar and edge feature points where unlike previous methods, the entire single cloud from the last scan goes through the process and not only the feature points. This approach allows fast registration of the points to the map while the accuracy increase by considering the entire cloud. Finally, two separate VoxelGrid filters (Han et al. 2017) are used to reduce the map size for planar and edge points. These filters create 3D grids (cubes) over the input point cloud data. Then, in each voxel, all the included points will be approximated and represented by the voxel centroid.

*4.6. Integration*:

The integration module takes the transforms computed in the transform estimation module and applies them to the map segments resulted from the feature registration module to form a continuous chain of map clouds with accurate pose change between any two scans. In other words, the poses of the lidar which has been computed gradually over the iterations of the transform estimation module are being fused into the map structure to have an accurate pose transform estimation in the final map. Any loop closure that has been detected during the mapping will be used here to correct the pose transform estimation.

*4.7. Loop Closure*:

To further eliminate drift, we have added a loop closure algorithm to detect previously visited locations. In such instances, new constraints are added if a match is found between the current feature set and a previous feature set using Iterative Closest Point (ICP) algorithm (Paul J. Besl and Neil D. McKay 1992). The estimated pose of the sensor is then updated by sending the pose estimates to the iSAM2 optimization algorithm (Kaess et al. 2012) which is a sparse incremental optimization method for nonlinear least-squares problems. iSAM2 algorithm works by incrementally estimating a set of variables (positions and landmarks in the environment) given a set of pose estimates. A detailed implementation of the iSAM2 algorithm for loop closure is presented in (Shan and Englot 2018). Figure 3 shows the detailed structure of the proposed SLAM system.

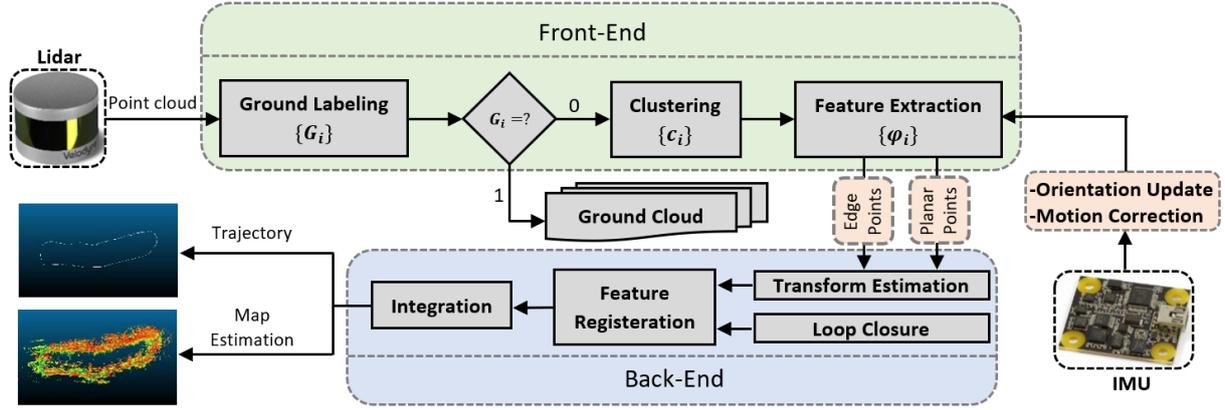

Figure 3. The full stack flowchart of the proposed Slam.

## 5. Experiments:

To test the performance of the proposed method, a set of experiments were conducted which will be described in the following sections.

*5.1. Hardware and software setup*

Two different hardware setups were used for data collection. First, a handheld device was used which was equipped with Velodyne VLP16 lidar, PhidgetSpatial Precision imu, a ZED stereo camera, and a computer with intel core i7 CPU and 32 GB of RAM running Ubuntu 18.04. The computer was the main processing source, and the camera feed was only used for visualization, and it was not included in the input data for the SLAM. Second, a mobile robotic platform (Superdroid HK1000-DM4-E) was used for collecting the data. The robot was equipped with Velodyne VLP16 lidar, PhidgetSpatial Precision imu, a Logitech C505 USB camera, and an onboard Jetson AGX Xavier with 32 GB of RAM and 8-Core ARM CPU running Ubuntu 18.04 for software implementation and data processing. The robot was also equipped with Septentrio AsteRx-U GNSS receiver with dual antenna. The GNSS readings were solely used for creating the ground truth data and evaluation purposes. The lidar has the frequency of 5Hz per sweep with 16 detector pairs, 360° horizontal field of view, and 30° vertical field of view with 600,000 points per scan. The maximum range of the lidar is 100(m) depending on the reflectivity of the targeted surface. Robot operating system (ROS) was used for sensor readings and recordings.

On both systems, sensor streams were published as ROS topics and a simple Python script was prepared to combine, synchronize, and record them into a single ROSBAG (.bag) file which is the standard recording format in ROS. Figure 4 shows the hardware setups. The ROS topics including lidar point cloud messages and imu messages were used for real-time implementation of the proposed SLAM system as well as for offline post-processing.

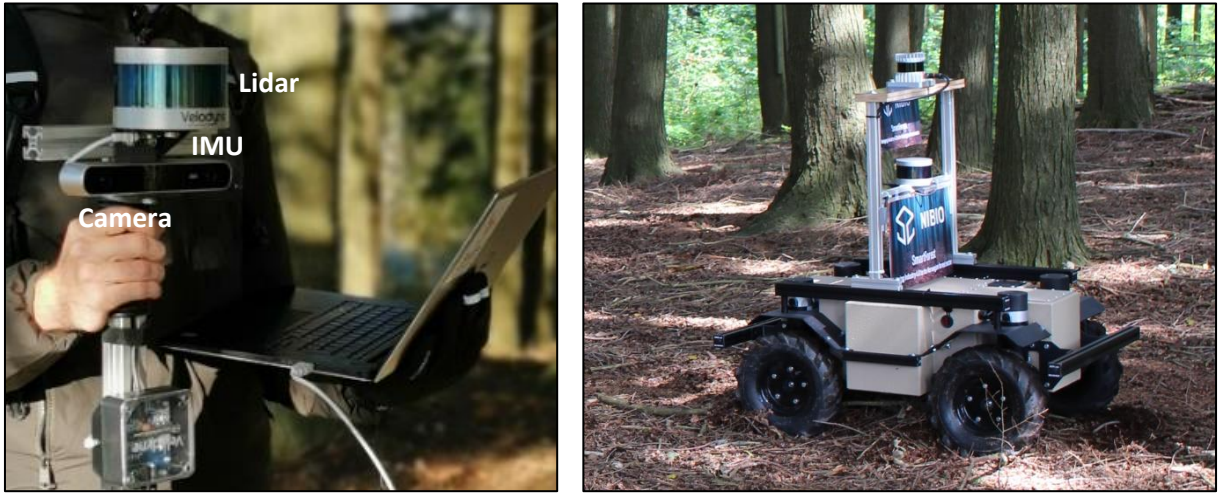

Figure 4. Data collection hardware. Left: handheld device, and right: mobile forestry robot.

## 5.2. Study area

Two forest stands were selected in the south of Oslo as shown in Figure 5. Both stands were selected based on number of trees, area, and the terrain. Since the maneuverability of the robot is greatly affected by the terrain characteristics, forests with extremely rough terrains were not considered for our data collection.

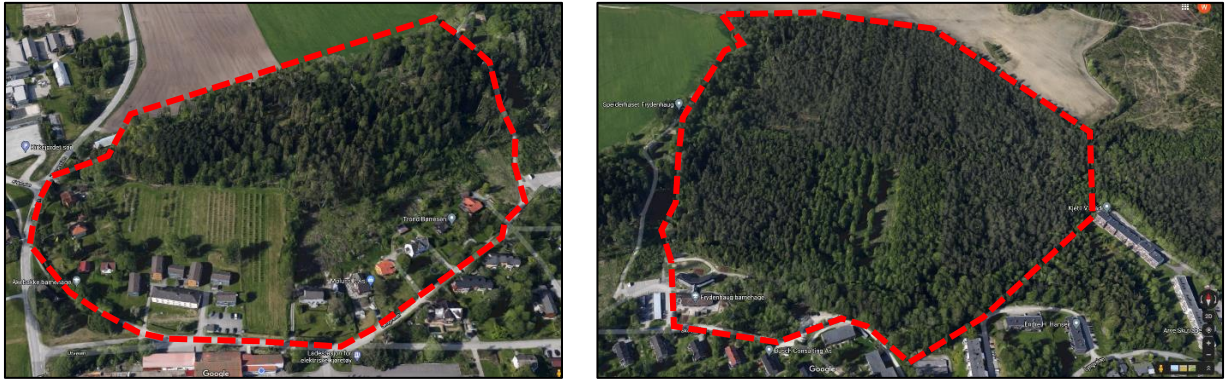

Figure 5. Two forest stands were selected for data collection.

## 5.3. Ground truth data

The performance of the proposed SLAM needs to be evaluated against ground truth data from the two forest stands used in data collection. This data includes the number of tree and DBH measurement. In stand 1, an area was selected, and number of trees were counted and the corresponding DBH were manually measured for each tree. In stand 2, two areas were selected and number of trees and DBH were measured for performance evaluation.

Table 1. Summary of the collected ground truth data.

| Area | Sample | # Trees | DBH (cm) | | | |
| --- | --- | --- | --- | --- | --- | --- |
| | | | avg. | std. | min. | max. |
| Stand 1 | 1.1 | 23 | 41.35 | 9.91 | 20.59 | 58.82 |
| Stand 2 | 2.1 | 26 | 34.34 | 13.43 | 18.51 | 61.48 |
| Stand 2 | 2.2 | 44 | 46.27 | 14.66 | 23.98 | 71.25 |

In all sections, only those trees which were completely inside the selected area were considered for the assessment. Table 1 summarized these data. These data were used to evaluate the proposed method in terms of number of trees, DBH estimations, and trajectory registration error.

*5.4.* Data collection

We have recorded two large datasets from stand 1 and stand 2. Total travelled distance was approximately 870(m) and 1320(m) for stand 1 and 2, respectively. The recorded data includes the lidar scans, imu readings, and camera feeds. A single frame of the lidar and camera recordings is shown in Figure X which was taken from stand 1 data. The data was recorded as ROSBAG packages which is ROS standard format and makes the sensor synchronization and integration easier.

The recorded data from stand 1 includes a complete loop to and from the forest. We have also made two small loops inside the forest to test the loop closure capabilities of the proposed SLAM. The handheld device was carried by the operator while walking from our main office to the forest and returning to the starting point. The walking speed was approximately 1(m/sec). This field test can be divided into three parts including walking in sideways while having buildings, cars, and small amount of vegetation around, walking in the forest with low density and nice distance between trees, and walking back to the starting point on the road with buildings and cars in the surroundings.

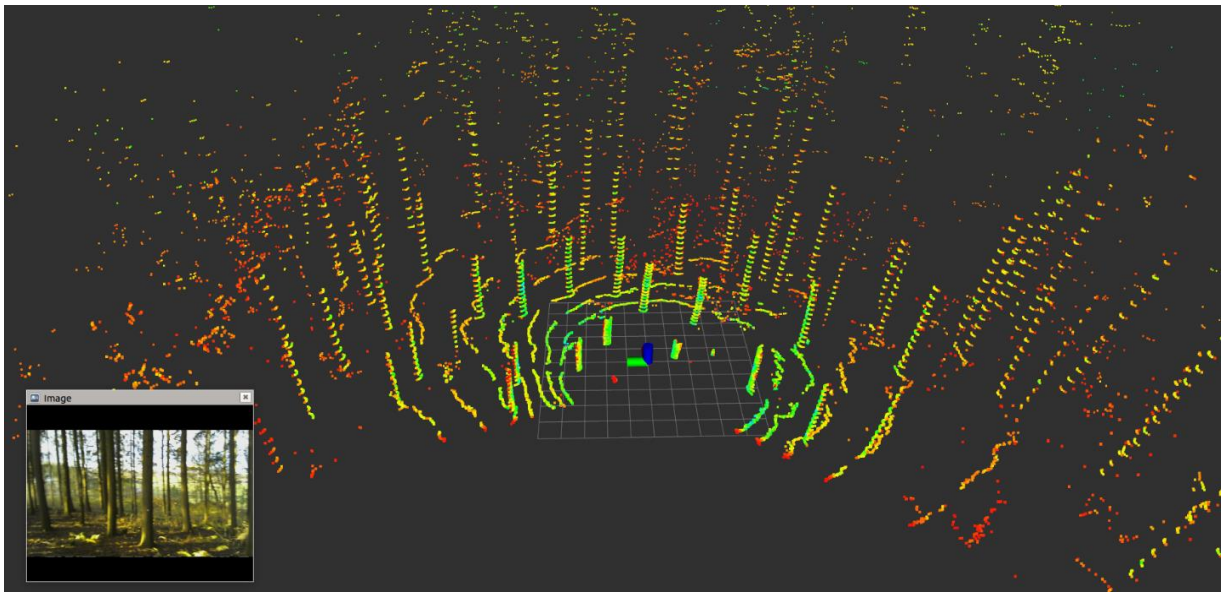

Figure 6. A frame from the recorded datasets including the lidar point cloud and the camera feed.

The recorded data from stand 1 includes a complete loop to and from the forest. We have also made two small loops inside the forest to test the loop closure capabilities of the proposed SLAM. The handheld device was carried by the operator while walking from our main office to the forest and returning to the starting point. The walking speed was approximately 1(m/sec). This field test can be divided into three parts including walking in sideways while having buildings, cars, and small amount of vegetation around, walking in the forest with low density and nice distance between trees, and walking back to the starting point on the road with buildings and cars in the surroundings.

For the second stand, we drove the robot from the entrance of the stand with three small loops and back to the starting position. The robot was mainly moving on a small forest dirt road in the middle of the stand and on three occasions was droved deep into the forest. The robot speed was set to two levels, including high speed of 2(m/sec) when driving on the dirt road, and low speed of 1(m/sec) when navigating in the forest. However, the robot faced several difficult parts of the forest terrain in which due to uneven surfaces, we had to drive it below the low speed or even stop it completely before turning around. In general, it was not possible to stabilize the speed of the robot. This is a common issue in forestry machines localization.

## 6. Results and Discussion:

The real-time implementation of the proposed SLAM system and the offline SLAM with the recorded ROSBAG data were both carried out and yield similar results. Therefore, the results are discussed based on the real-time implementation of the SLAM system. It should be noted that in the offline SLAM, it was possible to have uninterrupted performance with the playback rate of up to 3, which is an important advantage when dealing with large-scale datasets.

The real-time implementation of the proposed SLAM system was tested while recording the data and the resulting point cloud, tree models, and sensor trajectories were outputted. Table 2 summarizes the performance metrics in the test forest stands.

Table 2. Summary of the experiments.

| Area | Trajectory Length (m) | Elevation Change (m) | # Scans | Duration (sec) |
|---|---|---|---|---|
| Stand 1 | 873 | 8.5 | 13332 | 1344 |
| Stand 2 | 1728 | 3.5 | 55239 | 5571 |

*6.1.* Final Maps

The recorded data was continuously fed to the SLAM system to create a complete 3D map of the explored area, to register the location of the device over time, and to detect and measure the DBH of trees in the selected regions of the test environments. Figures 7 and 8 show the resulting maps, and the registered trajectories in stand 1 and 2 respectively where the results are visualized based on the vertical intensity and both maps were well-structured with expected loop closure corrections and tree models.

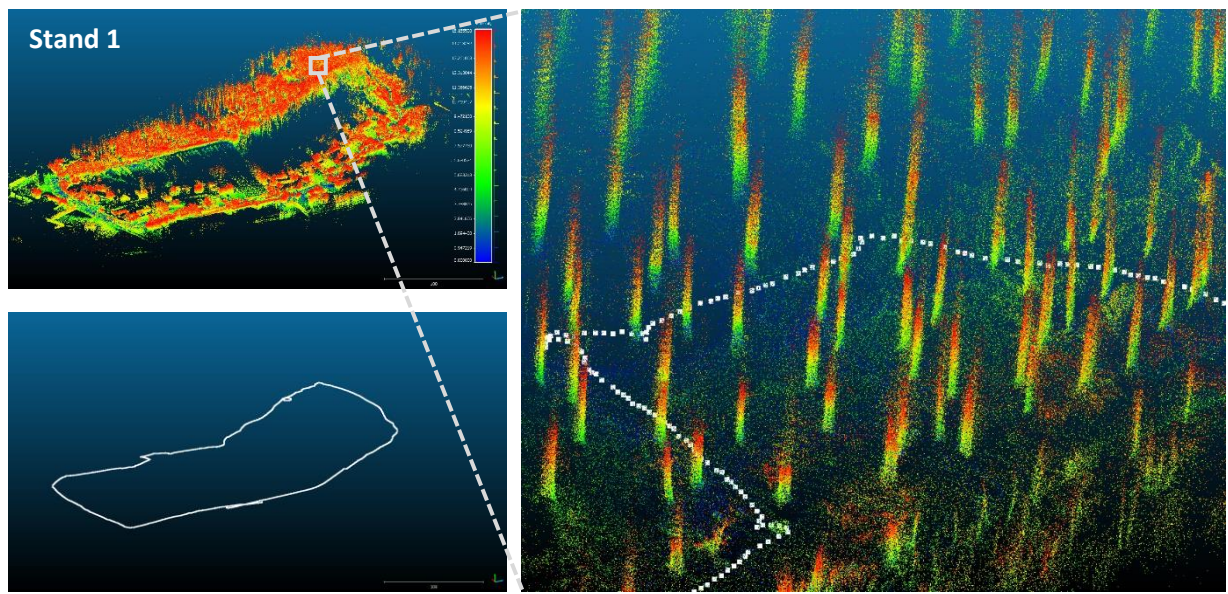

Figure 7. Resulted map in stand 1. Top left: final map, bottom left: corresponding trajectory, and right: a cut in the final map with visible trees, vegetations, and trajectory (white dotted line). The maps are colored based on vertical intensity.

The recorded data from stand 1 includes a complete loop to and from the forest. We have also made two small loops inside the forest and a big loop when returning to the initial position, to test the loop closure capabilities of the proposed SLAM system. The handheld device was carried by the operator while walking from our main office to the forest and returning to the starting point. The walking speed was approximately 1 (m/sec). This field test can be divided into three parts including walking in sideways while having buildings, cars, and small amount of vegetation around, walking in the forest with low density and nice distance between trees, and walking back to the starting point on the road with buildings and cars in the surroundings. The recording of stand 2 consists of the robot navigating from the entrance of the forest and explore the surrounding area and travelling back to the start location where, we have included four loop closures. Comparing to stand 1, we had a much denser forest with more vegetation. The robot speed was set to two levels, including high speed (2 m/sec) when driving on the dirt road, and low speed (1 m/sec) when navigating in the forest. However, the robot faced several difficult parts of the forest terrain in which due to uneven surfaces, we had to drive it below the low speed or even stop it completely before turning around.

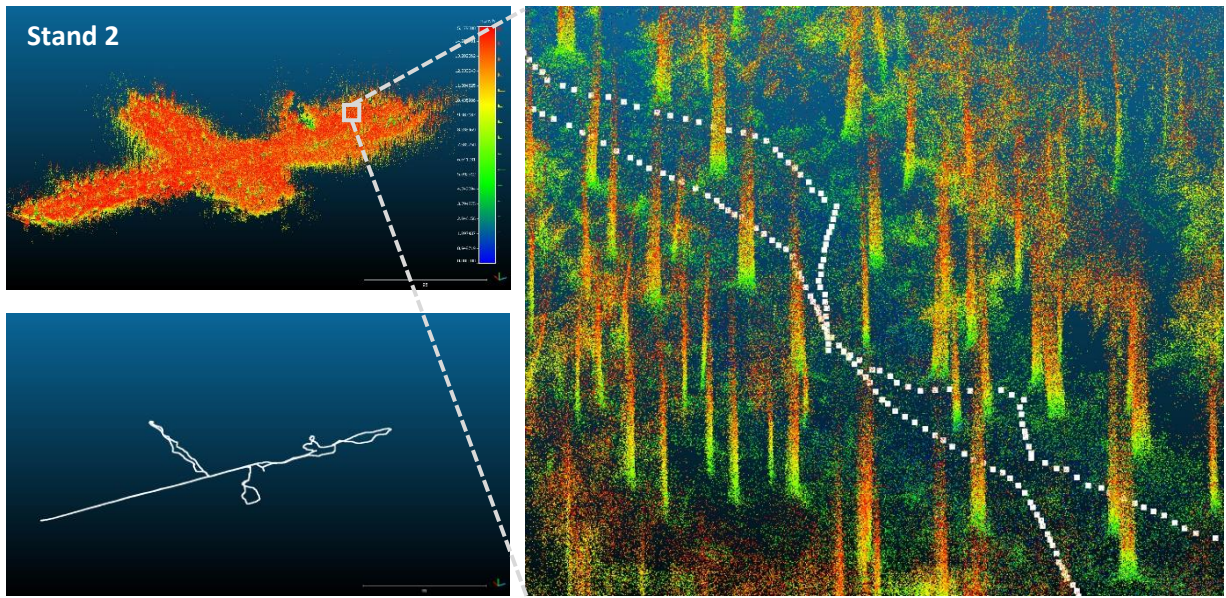

Figure 8. Resulted map in stand 2. Top left: final map, bottom left: corresponding trajectory, and right: a cut in the final map with visible trees, vegetations, and trajectory (white dotted line). The maps are colored based on vertical intensity.

The results show that the below-canopy structural information in the test plots is reconstructed by using the proposed SLAM system, and the trajectories are coincident with the practical movements. Each tree in the resulted structures is registered with no significant deviations, and the distribution of trees is identifiable, which suggests that the individual trees reconstructed by the proposed method have the potential to be applied to forest inventories. In addition, the trajectories of the mapping agent coincide with the relative tree locations and the visual assessment from the camera feed. However, since the implemented lidar has a limited vertical field of view and it was positioned to get more information from tree trunks and the ground points, the canopy is not fully registered in the resulted point cloud.

*6.2.* Tree analysis

As discussed in section 5.3, one section in stand 1 and two sections in stand 2 were selected for further performance analysis of our method. Figure 9 shows these three sections extracted from the resulted point clouds. The performance of the proposed SLAM system will be evaluated based on the performance in these three sections. To have a standard process of tree estimation, a minimum number of 100 points was required for each tree to be recognized. This is to detect only those trees that are completely inside each section since only those trees were included in the ground truth preparation phase.

At each assessment section, number of trees and the corresponding DBH estimations were computed and compared to the collected actual values described in section 5.3. Table 3 summarizes the estimated values and the actual values. The comparative results indicate an approximate average error of 3(cm) for DBH estimation. In sections 1.1 and 2.1, all trees were recognized and analyzed. However, two trees were missed in the final map of section 2.2. This inaccuracy is mainly because of the higher level of vegetation in section 2.2, smaller size of these two trees, and very short distance between them and neighbor trees which made the algorithm to cluster them as single trees with their neighbors.

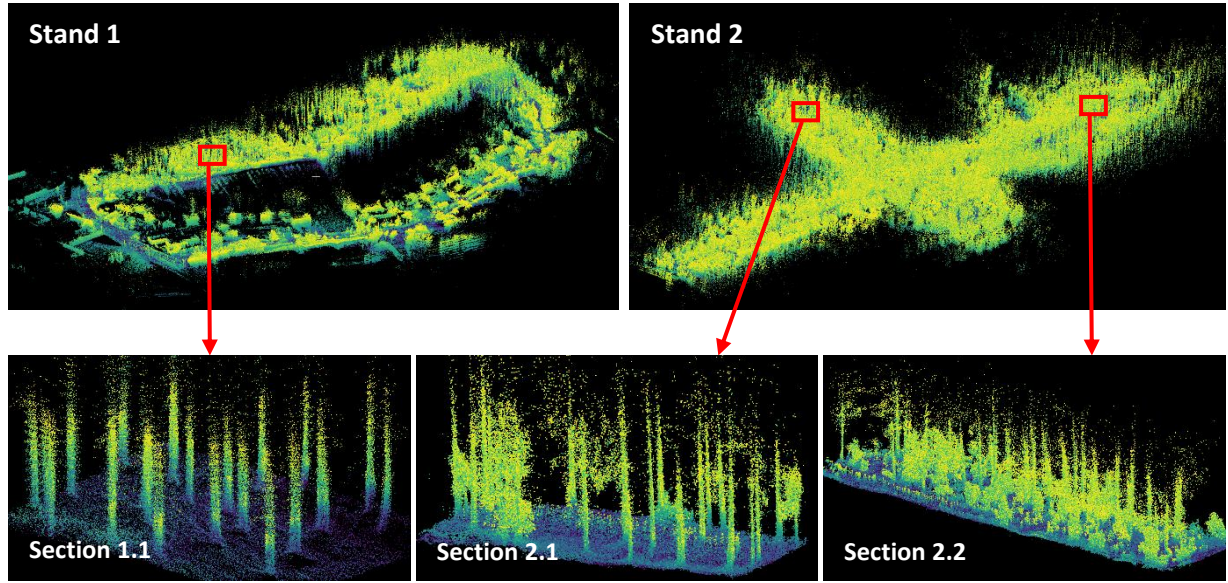

Figure 9. Corresponding point clouds of three selected sections from the experiment stands.

The resulting extracted tree models for these three sections are shown in Figure 10, where the ground points are removed, and single trees are visualized with different colors for better visibility. Detected trees are well-structured and the outlier removal component of the proposed system helps to have a more clean and distinguishable tree models. As explained before, the lidar placement is suitable for the trunk modelling and DBH estimation while the canopy and higher branches are not well-registered.

Table 3. Performance of the proposed SLAM system against the ground truth data from the three experimental sections in terms of number of trees and DBH (average, standard deviation, minimum, and maximum).

| Source | Stand 1 | | | Stand 2.1 | | | Stand 2.2 | | |
|---|---|---|---|---|---|---|---|---|---|
| | # Trees | DBH (cm) | | # Trees | DBH (cm) | | # Trees | DBH (cm) | |
| | | avg/std. | min/max | | avg/std. | min/max | | avg/std. | min/max |
| **Measured** | 23 | 41.35/9.91 | 20.59/58.82 | 28 | 34.34/13.43 | 18.51/61.48 | 44 | 46.27/14.66 | 23.98/71.25 |
| **Estimated** | 23 | 40.14/9.84 | 21.97/55.74 | 28 | 36.24/13.81 | 19.25/64.59 | 42 | 49.38/14.85 | 25.95/75.09 |

Since the ground points are detected and classified as a part of the SLAM process, DBH can be accurately estimated even in uneven regions where usual plane fitting methods might cause significant deviations. The outlier removal module improves the accuracy by detecting abnormalities in the point cloud mainly resulting from branches and leaves and sensor inaccuracies. Furthermore, resulting tree models are suitable for DBH estimation and further tree volume extraction as the entire trunk are visible and distinguishable from the final point cloud.

The DBH estimation error in these three sections are visualize in Figure 11, where estimated values are plotted against the ground truth data. In general, the estimation generated better results in section 1.1 of stand 1 since there was proper distances between individual trees and the amount of ground vegetation was much less than stand 2.

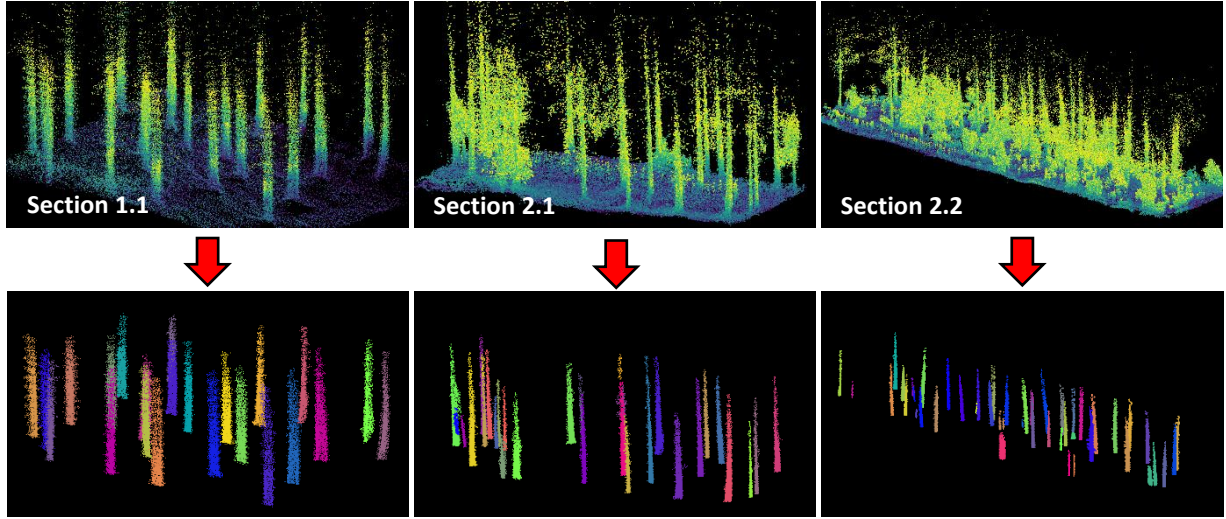

Figure 10. Extracted individual trees in each test section. Different colors are assigned to each detected tree to have a better visualization. Ground points and outlier points are removed.

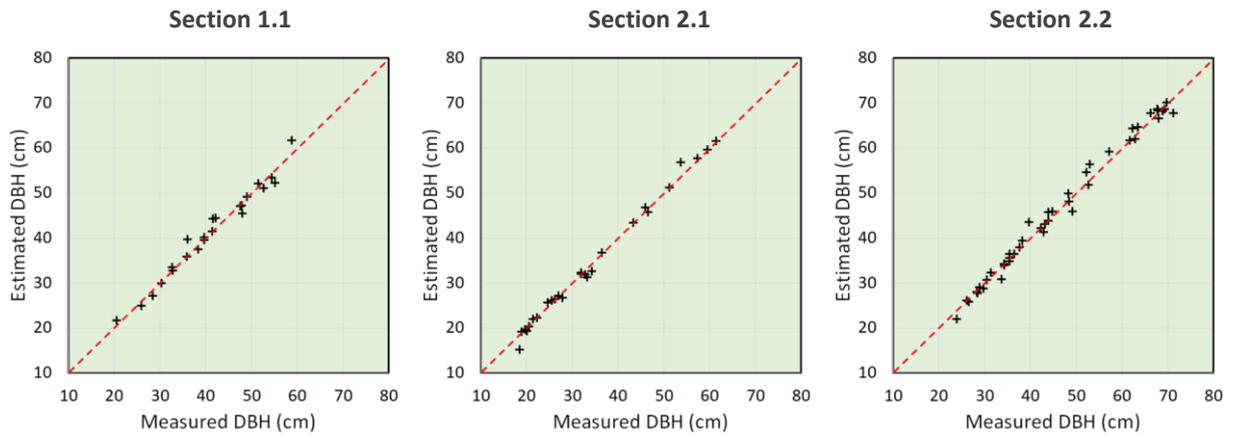

Figure 11. DBH estimation deviation from the ground truth data in three experimental sections.

Three error estimates were used to evaluate the resulted DBH values including Root Mean Square Error (RMSE), Mean Absolute Error (MAE), and Mean Absolute Percentage Error (MAPE) as formulated as follows:

$$RMSE = \left[ 1/N \left( \sum_{i=1}^{N}(x_i - y_i)^2 \right) \right]^{0.5},$$

$$MAE = 1/N \sum_{i=1}^{N} |x_i - y_i|,$$

$$MAPE = 1/N \sum_{i=1}^{N} \left| 1 - \frac{x_i}{y_i} \right|,$$

where $x_i$, $y_i$, and $N$ are predicted values, actual values, and number of samples, respectively. Table 4 shows the error values for the proposed SLAM system when estimating DBH in test sections.

DBH estimates are more accurate in section 1.1 while section 2.2 resulted in higher error values. MAE value of 3.98 (cm) in section 2.2 shows the effect of forest density and vegetation on DBH estimation. When obtaining 3D point cloud data in forest stands with higher density trees or undergrowth vegetation, it is more likely that trees will block one another and undergrowth vegetation will block the trunks, producing incomplete point cloud data at a certain height which leads to missing or misidentification of trees.

Table 4. Error estimates for the resulting DBH values in test sections.

| DBH Estimation Error | RMSE (cm) | MAE (cm) | MAPE (%) |
|---|---|---|---|
| Section 1 | 2.31 | 1.89 | 3.12 |
| Section 2.1 | 2.79 | 2.47 | 3.83 |
| Section 2.2 | 3.52 | 3.98 | 4.76 |

However, the performance of the proposed SLAM system is still highly reliable considering the cost and resolution of the used sensors and the data collection time. Using a lidar with higher resolution and spending many hours recording the stand from different angels might improve the accuracy but won't be practical considering the total cost of the process.

An illustration of the estimated DBH values is shown in Figure 13 where in each test section, top view of the scene is provided with a circle representing each tree with the diameter equal to the estimated DBH for that tree. In sections 1.1 and 2.1, all trees were detected from the resulting point cloud and only two trees were not detected in section 2.2 because they were both small, and very close to one of their neighbor trees.

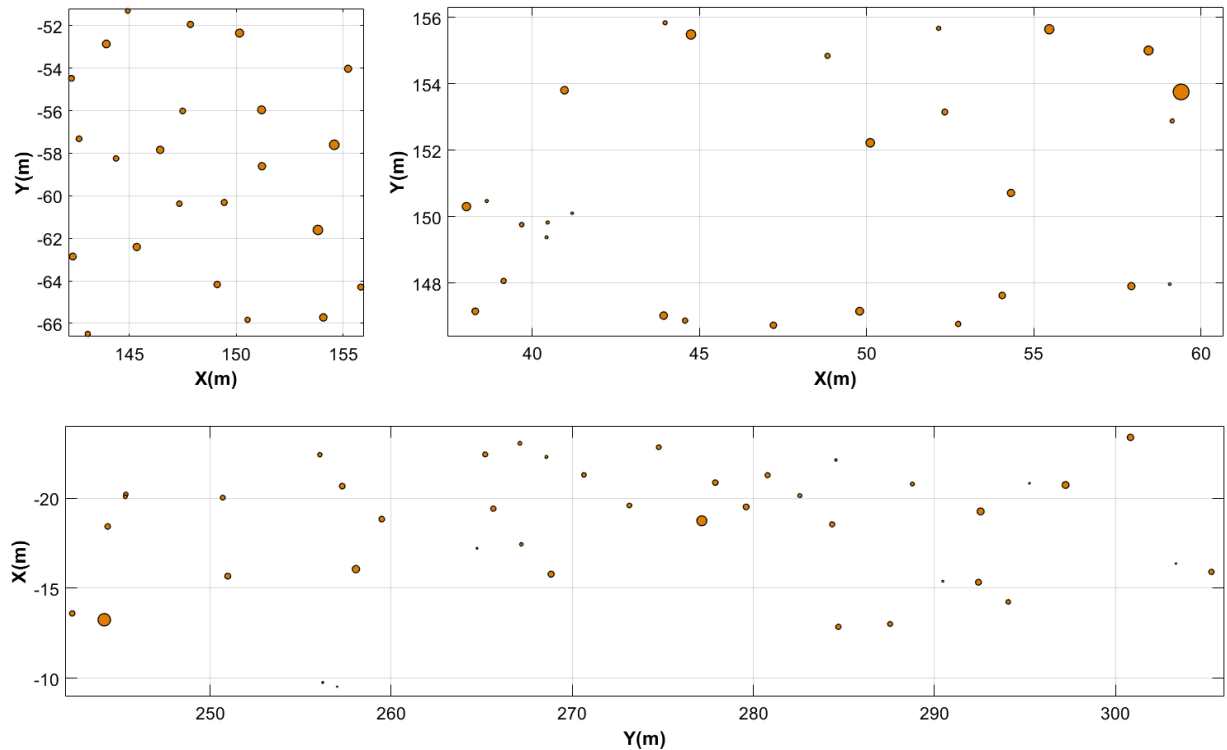

Figure 12. 2D illustration of the extracted DBH values in test sections.

### 6.3. Loop closure

The performance of the proposed SLAM was further improved by including loop closure modules. The proposed ICP-based loop closure was implemented with iSAM2 optimization. Figure 14 shows an instance of loop closure in stand 1 were the robot returns to the starting position and the trajectory and the corresponding map were adjusted. We found adding loop closure module is especially useful for correcting the drift in a sensor's altitude. In Figure 13, the inaccuracy in the altitude is visible before the loop closure and it is corrected as the handheld device get closer to the initial position. Even though it is much smaller than the altitude drift, the drift in the $x$ and $y$ coordination is also corrected by performing loop closure instances. The ability to perform loop closures in the online implementation of the proposed SLAM system makes it a useful tool for large-scale mapping tasks.

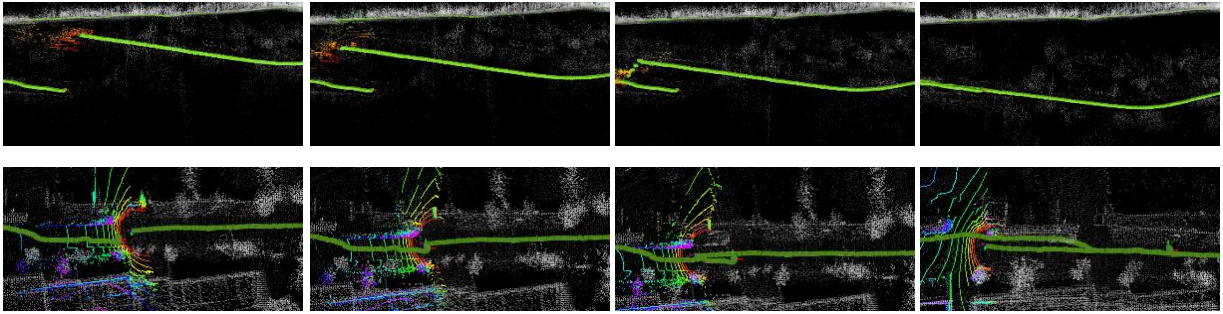

Figure 13. An instance of loop closure implementation in the proposed SLAM system. From left to right, top: 3D view of the trajectory and resulting map being corrected as the Lidar gets close to the original position, and bottom: top view of the same frames with the Lidar scans and the trajectory.

*6.4.* Comparative studies:

To further evaluate the functionality of the proposed method, a set of 3 benchmark SLAM algorithms were selected for comparative studies including LOAM (Zhang and Singh 2017), LeGO_LOAM (Shan and Englot 2018), and Hdl_graph_slam (Koide et al. 2019b). Both recorded datasets were fed to each algorithm to compare with the proposed SLAM system.

The first evaluation is on the pose estimation error where we compare the starting pose of the sensor with the final pose when it returns to the initial location. In both data recording instances, the final pose was set carefully to match the starting pose. Therefore, we could compute the rotational (roll, pitch, yaw) and translational (x, y, z) errors as presented in Table 5.

Table 5. Pose estimate error when returning to the start location.

| Test | Method | Roll (º) | Pitch (º) | Yaw (º) | Total Rot. (º) | X (m) | Y (m) | Z (m) | Total Trans. (m) |
|---|---|---|---|---|---|---|---|---|---|
| Stand 1 | LOAM | 8.56 | 6.15 | 9.83 | 14.41 | 1.65 | 1.98 | 2.12 | 3.34 |
|  | LeGo_LOAM | 7.62 | 4.33 | 7.65 | 11.63 | 0.68 | 0.42 | 1.61 | 1.80 |
|  | Hdl_graph_slam | 8.97 | 5.14 | 7.67 | 12.87 | 0.89 | 0.66 | 1.83 | 2.14 |
|  | Proposed method | 5.68 | 3.95 | 6.14 | 9.25 | 0.48 | 0.33 | 1.54 | 1.65 |
| Stand 2 | LOAM | 12.38 | 11.17 | 15.63 | 22.85 | 8.34 | 9.08 | 12.38 | 17.47 |
|  | LeGo_LOAM | 9.88 | 9.84 | 12.56 | 18.77 | 3.69 | 4.33 | 3.38 | 6.62 |
|  | Hdl_graph_slam | 10.69 | 9.15 | 11.45 | 18.14 | 4.71 | 5.1 | 2.75 | 7.47 |
|  | Proposed method | 7.38 | 6.55 | 11.68 | 15.29 | 2.73 | 3.91 | 3.22 | 5.75 |

The proposed method yields superior results with 18% improvement in total rotation estimation and 11% improvement in total translation estimation averaged over both stands.

The second evaluation is based on comparing motion estimations. Since we registered the position of the robot in stand 2 with the onboard GNSS device, the distance between a point on the SLAM trajectory and the GNSS readings was calculated at time t with the frequency of 0.1 Hz which equals to one point every 10 seconds. Table 6 shows the average values for the motion estimation drifts in stand 2. The comparative results show an average improvement of 17% in motion estimation accuracy.

Table 6. Relative error for motion estimation. These values are obtained from 557 points in stand 2 which estimates to one point per 10 seconds.

| Test | Method | Mean (m) | Std. | Min (m) | Max (m) |
|---|---|---|---|---|---|
| Stand 2 | LOAM | 6.72 | 2.97 | 1.55 | 7.65 |
|  | LeGo_LOAM | 4.68 | 1.65 | 0.15 | 6.28 |
|  | Hdl_graph_slam | 5.29 | 2.33 | 0.35 | 7.36 |
|  | Proposed method | 3.87 | 1.02 | 0.05 | 5.84 |

## 7. Conclusion:

In this study, we have explored the integration of robotics and sensor fusion techniques to address the challenging task of real-time forest inventory in large-scale, GNSS-interrupted forest environments. Forested environments, with their dense canopies and challenging terrain, often disrupt GNSS signals. Through the utilization of sensor fusion techniques, such as LiDAR and IMU, our proposed method has demonstrated the capacity to maintain accurate localization and mapping capabilities even in areas with limited GNSS reception. This feature is important in enabling continuous data collection for forest inventory, regardless of the presence of GNSS signals. Also, the integration of robots into the forest inventory process significantly enhances efficiency which is vital for keeping forest inventory data up-to-date and comprehensive.

While this research marks a step forward in real-time forest inventory, there are still avenues for further exploration. Future work should focus on refining sensor technologies, expanding the scope to include additional environmental factors, and enhancing the autonomy and adaptability of robotic platforms in dynamic forest environments.

## 8. Acknowledgement: